\documentclass[letterpaper, 10 pt, journal, twoside]{IEEEtran}

\ifCLASSINFOpdf
\else
\fi

\usepackage{amsmath,amsfonts}
\usepackage{algorithmic}
\usepackage{algorithm}
\usepackage{array}
\usepackage[caption=false,font=normalsize,labelfont=sf,textfont=sf]{subfig}
\usepackage{textcomp}
\usepackage{stfloats}
\usepackage{url}
\usepackage{verbatim}
\usepackage{graphicx}
\usepackage{cite}
\usepackage{overpic}
\usepackage{xcolor}
\usepackage{booktabs}
\usepackage{multirow}

\definecolor{revisedorange}{HTML}{FF8000}
\newcommand{\revised}[1]{#1}

\definecolor{myBlue}{HTML}{0000FF}
\definecolor{myPink}{HTML}{FF69B4}
\definecolor{myCyan}{HTML}{00B3B3}
\definecolor{lethalred}{HTML}{8B2E16}
\definecolor{goalgreen}{HTML}{A3B18A}

\begin{document}

\title{A Terrain-Adaptive $\epsilon$-Constraint MPC for Uneven Terrain Kinodynamic Planning}

\author{Otobong~Jerome$^{1}$, Geesara~Kulathunga$^{2}$, and Tiago~Nascimento$^{1}$%
\thanks{Manuscript received: March 1, 2026; Revised: May 20, 2026; Accepted: July 6, 2026.}%
\thanks{This paper was recommended for publication by Editor Aniket Bera upon evaluation of the Associate Editor and Reviewers' comments.
This work was supported by the National Council for Scientific and Technological Development (CNPq) under research project No.~401097/2025-0, and by Petrobras under the cooperation agreement project No.~0050.0132311.25.9.}%
\thanks{$^{1}$Otobong Jerome and Tiago Nascimento are with the Laboratório de Engenharia de Sistemas e Robótica, Universidade Federal da Paraíba (UFPB), João Pessoa, Paraíba, 58055-000, Brazil.
{\tt\small \{o.jerome@laser.ci, tiagopn@ci\}.ufpb.br}}%
\thanks{$^{2}$Geesara Kulathunga is with the School of Computer Science, University of Lincoln, Lincoln, United Kingdom.
{\tt\small gkulathunga@lincoln.ac.uk}}%
\thanks{Digital Object Identifier (DOI): see top of this page.}%
}

% Paper headers
\markboth{IEEE Robotics and Automation Letters. Preprint Version. Accepted July, 2026}%
{Jerome \MakeLowercase{\textit{et al.}}: A Terrain-Adaptive $\epsilon$-Constraint MPC for Uneven Terrain Kinodynamic Planning}

\maketitle

\begin{abstract}
Kinodynamic planning for car-like vehicles on uneven terrain requires \revised{jointly optimising objectives such as path efficiency and pose stability, which often conflict on challenging terrain}. This work formulates the problem as a multi-objective optimisation and solves it using the $\epsilon$-constraint method, a classical technique that minimises one objective while constraining each remaining objective to lie below an upper bound $\epsilon$. The method is embedded in a Model Predictive Control (MPC) framework in which the $\epsilon$ bounds are adjusted online from local terrain descriptors, allowing the planner to explore the Pareto front in real time. The motion of the vehicle over the terrain is captured by a semi-parametric model that
combines analytical car-like dynamics with a Sparse Gaussian Process trained on the same terrain descriptors. The proposed $\epsilon$-MPC is evaluated against two sampling-based baselines, Model Predictive Path Integral (MPPI) control and Genetic-Algorithm Kinodynamic (GAKD)
planning, achieving a $94\%$ navigation success rate while reducing maximum orientation deviation by up to $24\%$ and improving multi-objective trade-off quality by up to $23\%$ relative to MPPI.
\end{abstract}


\section{Introduction}
\IEEEPARstart{V}{arious} methods generate triangular meshes from sensor data~\cite{whelan2015real}, including large-scale environments. Robotics frameworks like Move Base Flex (MBF)~\cite{putz2018move} provide adaptable tools for building so-called Navigation Meshes and enabling robot navigation. Yet navigation remains challenging for autonomous robots, especially in complex unstructured terrains such as construction sites, disaster zones, or planetary surfaces. Robust navigation there demands planners that compute collision-free paths while respecting the kinodynamic constraints of the robot's dynamics (velocity and acceleration limits); however, accurate vehicle-terrain modelling is itself difficult due to terrain effects.

Traditional mesh planners such as the continuous vector planner (CVP) \cite{putz2021continuous} apply the Fast Marching Method (FMM) \cite{sethian1996fast} on multilayer 3D triangular meshes to generate smooth geodesic paths over rough terrain, yielding a global, continuous shortest-path vector field that encodes terrain attributes such as roughness and elevation. However, they do not explicitly incorporate kinodynamic constraints, limiting their use where dynamic feasibility is critical.

Kinodynamic planning addresses motion planning under both kinematic and dynamic constraints. The first polynomial-time, provably good approximation, with complexity analysis, was given in \cite{johncanny1988}, while randomised approaches such as RRT and RRT\* variants incorporate kinodynamic cost optimisation \cite{L4}. Recent work splits into two categories: non-data-driven methods, where gradient-free schemes like Model Predictive Path Integral (MPPI) control handle complex dynamics without explicit gradients; and data-driven methods, which use models such as Recurrent Neural Networks (RNN) and Deep Reinforcement Learning (DRL) to learn dynamics and optimise planning in high-dimensional or partially observed settings.

Recent DRL advances show promise for learning navigation policies in dynamic environments \cite{RL1}, yet key limitations persist: most approaches are confined to 2D planning with sparse rewards and generalise poorly across start-goal configurations, lack integration with continuous global guidance, and ignore kinodynamic constraints, particularly on uneven terrain. Standard RL rewards also fail to capture the multi-objective nature of the problem, where path optimality and physical stability must be jointly considered. The $\epsilon$-constraint method addresses this: \revised{a classical multi-objective technique that minimises one primary objective while constraining each of the remaining objectives to lie below an upper bound $\epsilon$; sweeping these bounds over a range of values recovers the Pareto front. In this work, the path-length objective is minimised, while the pose-stability objective is constrained by a terrain-dependent upper bound, denoted $\epsilon_{\text{stab}}$ throughout the paper.} Although widely applied elsewhere, $\epsilon$-constraint methods remain under-explored for uneven terrain kinodynamic planning. Hence, the contributions of this work are:

\begin{enumerate}
    \item \revised{A semi-parametric vehicle-terrain model in which a Sparse Gaussian Process corrects a nominal kinematic model, reducing roll and pitch prediction error by up to $32\%$. The process is conditioned on a novel local terrain descriptor combining surface normal, roughness, neighbourhood extent, and Gaussian and mean curvatures, the latter obtained by extending per-vertex discrete operators to arbitrary query points via distance-weighted interpolation.}
    \item A terrain-adaptive $\epsilon$-constraint reformulation of the bi-objective kinodynamic planning problem, in which the stability tolerance $\epsilon_{\text{stab}}$ is scheduled online from local roughness and slope within the receding-horizon loop.
    \item An FMM-biased sampling strategy with feasibility selection under the adaptive $\epsilon_{\text{stab}}$, integrating global guidance, local exploration, and terrain-aware constraint enforcement in a single real-time planner.
    \item An open-source ROS implementation compatible with Move Base Flex, evaluated in simulation and on two physical platforms across uneven outdoor terrain.
\end{enumerate}

\section{Related Work}
\begin{figure*}[!ht]
  \centering
  \begin{overpic}[width=0.6\linewidth]{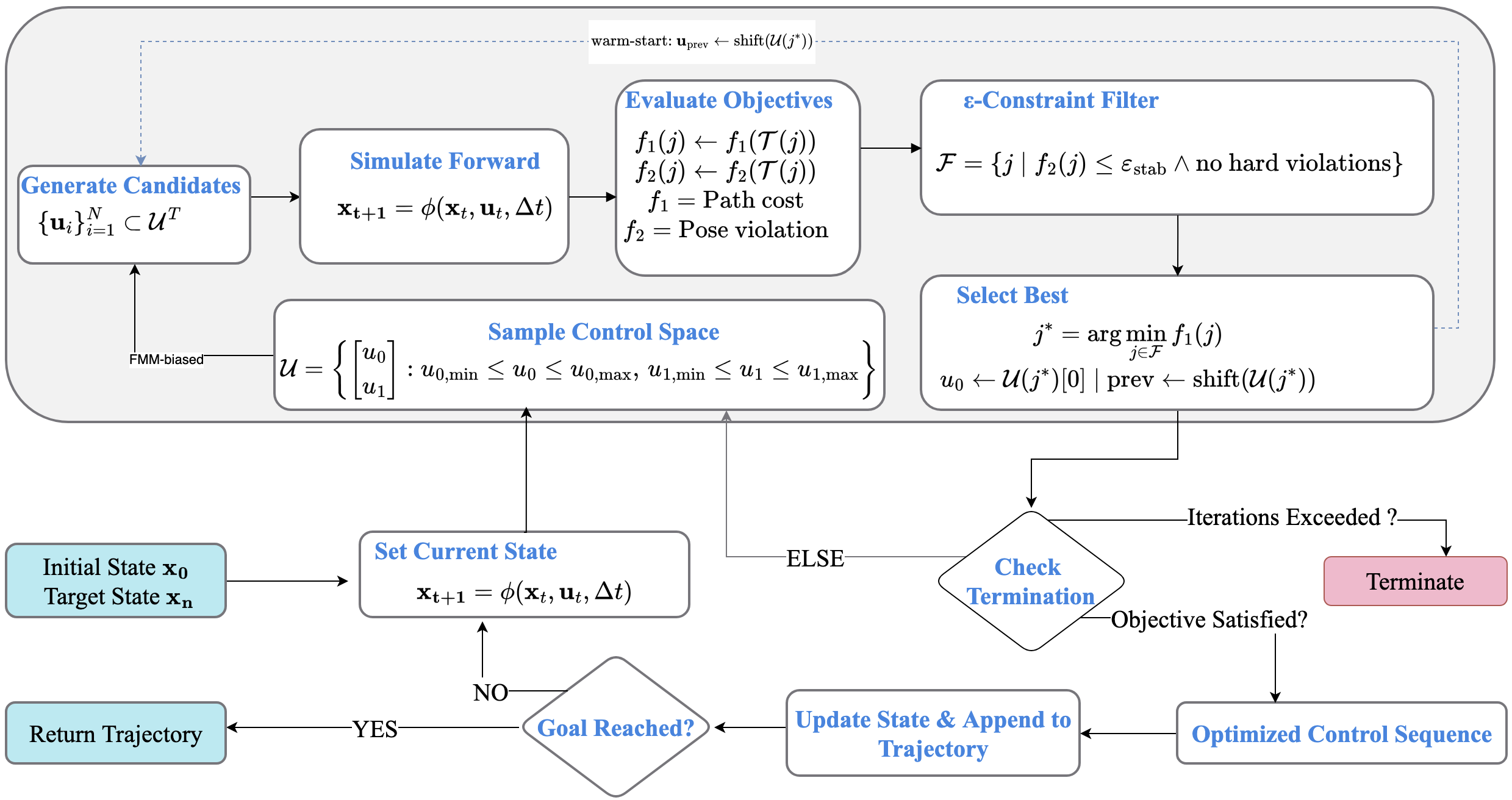}
  \end{overpic}
  \caption{Flow chart of the proposed kinodynamic planner: A terrain-adaptive $\epsilon$-constraint optimisation is used to find the sequence of controls that best satisfies both objectives $(f_1 ,f_2)$ while ensuring all the controls in the sequence are from the space of valid controls.}
  \label{fig:abstract}
\end{figure*}
Kinodynamic planning on 3D triangular meshes draws heavily on geodesic computation. Classic discrete methods such as MMP~\cite{mitchell1987discrete} and CH~\cite{chen1990shortest} (and their variants) are precise on smaller meshes, while the fast marching method (FMM)~\cite{sethian1996fast} and heat method~\cite{crane2013heat} scale better to large datasets; the more recent edge-flip method~\cite{crane2020flip} refines paths via local mesh modifications at higher computational cost (surveys: \cite{bose2011survey,mikd}). Planning on such meshes then balances proximity to the optimal geodesic against the robot's kinematic and dynamic constraints.

Optimisation-based methods such as Sequential Quadratic Programming (SQP) \cite{Fletcher2010}, Model Predictive Control (MPC) \cite{pari2013model}, Covariant Hamiltonian Optimization for Motion Planning (CHOMP) \cite{zucker2013chomp}, and Iterative Linear Quadratic Regulator (ILQR) converge well but are sensitive to initialisation, prone to local optima, and costly online, particularly in non-linear, non-convex settings. Sampling-based alternatives like Model Predictive Path Integral Control (MPPI) \cite{williams2016aggressive} and Genetic-Algorithm Kinodynamic (GAKD) planning \cite{jerome2025} avoid gradients entirely: MPPI evaluates many independent control sequences in parallel for broad coverage, while GAKD instead recombines the best trajectory segments via a genetic algorithm, giving robust real-time performance in challenging environments. Both, however, rely on accurate dynamics models that are hard to derive analytically for car-like vehicles on arbitrary 3D mesh terrain; one contribution of this paper is a semi-parametric approach blending data-driven flexibility with physical structure. In MPPI and GAKD, weighted scalarisation suffers a structural scale mismatch: path length accrues cost at every step, while the stability term activates only on steps violating the safe band. With fixed weights, the sum is dominated by path length wherever stability is zero, and no single weight balances the trade-off across terrains. The $\epsilon$-constraint reformulation instead treats stability as a constraint, so the planner minimises path length freely when the constraint is slack and backs off only when it is active.

The above assumes a static, offline-constructed mesh. A complementary line estimates traversability online from onboard sensors, via implicit neural representations queried at arbitrary locations~\cite{TRAIL2026}, terrain-aware policies from RGB images and vehicle pose~\cite{Jia2025}, and physics-informed evidential learning for out-of-distribution terrain~\cite{PIETRA2025}. These are orthogonal to our work: they address where the terrain model comes from, whereas we address how to plan given one.

Reinforcement Learning (RL) shows promise for mobile robot navigation but suffers from sparse rewards, slow convergence, poor generalisability, and limited interpretability. Hybrid strategies address these: hierarchical architectures pairing global planners with local RL agents \cite{RL10}, sensor-based feedback and transfer learning \cite{RL9}, modular designs decoupling deep learning components \cite{RL8}, and attention mechanisms for interpretability \cite{RL1} (reviewed in \cite{review2}). However, these are mostly designed for 2D planar environments or rely on elevation maps that fail to capture the intricacies of uneven 3D mesh terrain.

Hence, a planner is needed that explicitly addresses the multi-objective nature of kinodynamic planning, balancing path optimality against physical stability while generalising to unseen uneven terrain.

The $\epsilon$-constraint method is a classical multi-objective technique; for example, \cite{nosratabadi2024intelligent} applied it to intelligent parking-lot power management and engineering optimisation. Though effective at generating Pareto-optimal solutions, its computational cost limits practicality in high-dimensional or real-time settings.

We circumvent this by sampling from the Pareto front within a receding horizon, with the stability constraint adaptively tightened or relaxed by local terrain conditions. A scaled continuous vector field provides global guidance for consistent performance across diverse meshes, the terrain-adaptive $\epsilon$-constraint explores the trade-off frontier between \revised{the two objectives}, and the semi-parametric vehicle-terrain model enables offline development and online generalisation.

\section{Preliminaries}\label{sec:preliminaries}
  \subsection{Discrete-time Vehicle Kinematics}
The control inputs are a linear velocity $u_0$ along the vehicle's forward $x$-axis and an angular velocity $u_1$ about the vertical $z$-axis, applied for duration $\Delta t$. The state and input are $
\mathbf{x} = \begin{bmatrix} x & y & z & \phi & \theta & \psi \end{bmatrix}^{\top}$, $\mathbf{u} = \begin{bmatrix} u_0 & u_1 \end{bmatrix}^{\top}$, with position $\mathbf{p} = \begin{bmatrix} x & y & z \end{bmatrix}^{\top}$ and roll-pitch-yaw $\boldsymbol{\Theta} = \begin{bmatrix} \phi & \theta & \psi \end{bmatrix}^{\top}$.

\subsection{Fast Marching Method on Triangle Meshes}
The Fast Marching Method (FMM) \cite{sethian1996fast}, as employed in \cite{putz2021continuous}, computes a continuous vector field over a triangle mesh in $O(n \log n)$ time ($n$ = passable vertices), accessible anywhere by barycentric interpolation: for direction vectors $\mathbf{d}_A, \mathbf{d}_B, \mathbf{d}_C$ at vertices $A, B, C$ and barycentric coordinates $(\lambda_1, \lambda_2, \lambda_3)$ of a query point $\mathbf{p} \in \triangle ABC$, $\mathbf{d}_P = \lambda_1 \mathbf{d}_A + \lambda_2 \mathbf{d}_B + \lambda_3 \mathbf{d}_C$ \cite{putz2021continuous}. The vertex costs incorporate height differences, roughness, border penalties, and inflation zones, so the field naturally encodes obstacle avoidance and terrain traversability \cite{putz2018move, putz2021continuous}. 

\begin{figure}[htbp]
    \centering
    \begin{overpic}[width=0.6\linewidth]{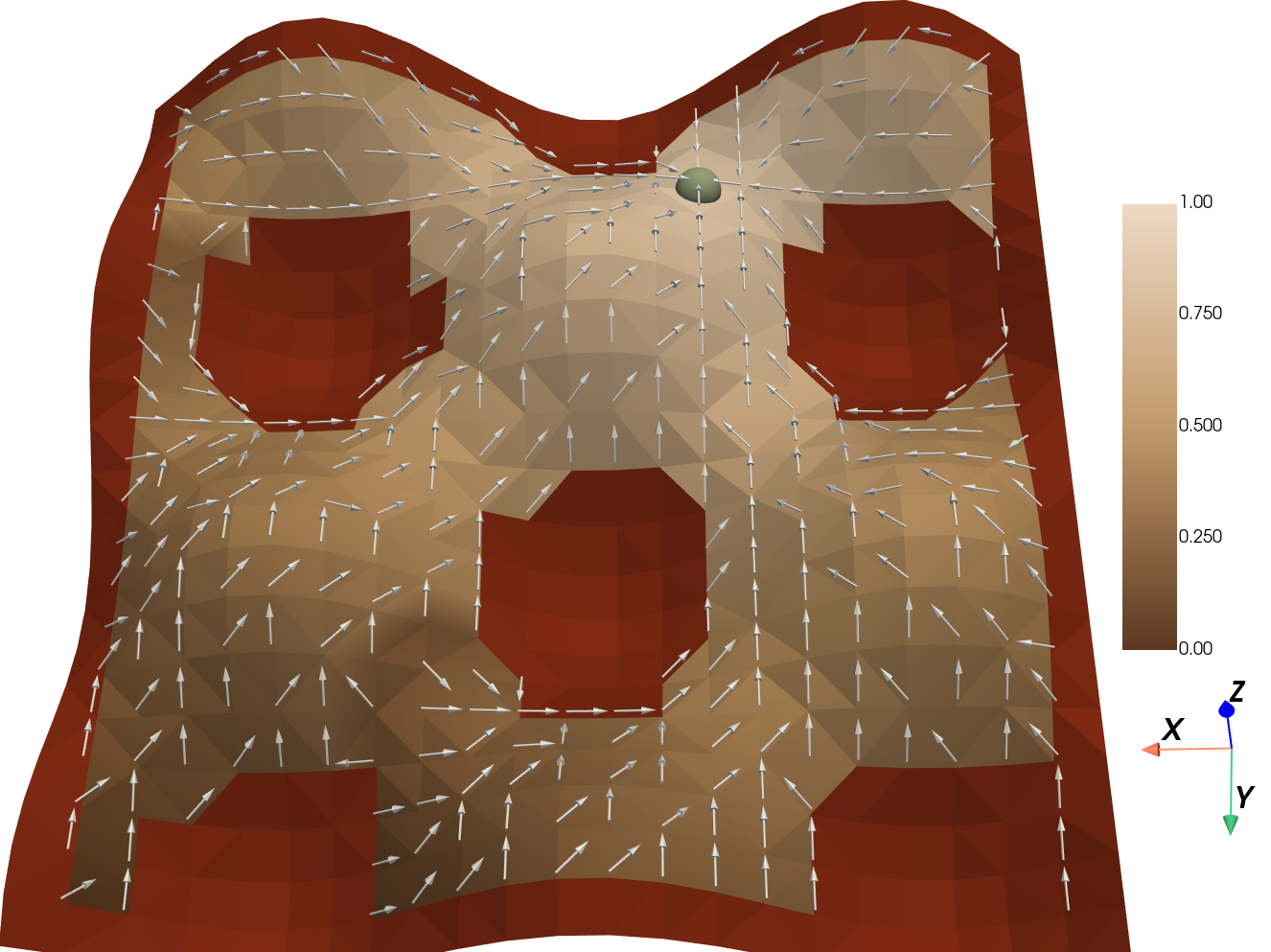}
        \put(5,74){\textcolor{goalgreen}{\Large\textbullet\ Goal}}
        \put(41,74){\textcolor{lethalred}{\Large\textbullet\ Lethal}}
        \put(87,62){$s(\mathbf{p})$}
    \end{overpic}
    \caption{Vector field over a mesh guiding an agent toward the goal while avoiding lethal regions. The colour gradient represents the scalar scaling function $s(\mathbf{p})$ ($\mathfrak{s} = 2$), with intensity increasing near the goal.}
    \label{fig:sp}
\end{figure}

    \subsection{Discrete Differential Geometry Operators}
     The discrete Gaussian curvature at vertex $v$ is computed via the angle-defect formula~\cite{Meyer2003} $K_v = (2\pi - \sum_{j=1}^{\mathfrak{f}} \theta_j)/A_{\text{Mixed}}^v$, where $\theta_j$ are the face angles at $v$ and $A_{\text{Mixed}}^v$ is the mixed Voronoi area~\cite{Meyer2003}. The discrete mean curvature is obtained from the Laplace-Beltrami operator
    $\boldsymbol{\kappa}(\mathbf{p}_v) = \frac{1}{2A_{\text{Mixed}}^v} \sum_{u \in \mathfrak{N}(v)} (\cot \alpha_{vu} + \cot \beta_{vu}) (\mathbf{p}_v - \mathbf{p}_u)$, yielding $H_v = \frac{1}{2} \| \boldsymbol{\kappa}(\mathbf{p}_v) \|$, where $\alpha_{vu}$ and $\beta_{vu}$ are the angles opposite edge $(v,u)$ in adjacent triangles. The mixed area $A_{\text{Mixed}}^v$ ensures robust computation: for non-obtuse triangles, the Voronoi region area is used, while for obtuse triangles, the appropriate fraction of the triangle area is substituted.

\section{Problem Formulation}
The system is described by: $\mathbf{x}_{i+1} = f(\mathbf{x}_i, \mathbf{u}_i, \mathfrak{d}_i, \Delta t)$ with $\mathbf{x}_0 = \mathbf{x}_{\text{start}}$ and goal $\mathbf{p}_{\text{goal}}$, where $\mathfrak{d}_i$ is a local terrain descriptor characterising the mesh geometry in a neighbourhood of $\mathbf{p}_i$, defined in Section~\ref{sec:descriptor}. At each planning iteration $k$, a multi-objective optimisation is solved over horizon $H$ from the current state $\mathbf{x}_k$, producing an optimised trajectory: $\mathcal{T}_k = \{\mathbf{x}_i^k \mid i=0,\ldots,H\} \cup \{\mathbf{u}_i^k \mid i=0,\ldots,H-1\}, \quad \mathbf{x}_0^k = \mathbf{x}_k.$ The first control action is then executed, the state is advanced to $\mathbf{x}_{k+1} = \mathbf{x}_1^k$, and the plan is recomputed. This process is repeated until $\|\mathbf{p}_k - \mathbf{p}_{\text{goal}}\| < \delta_{\text{term}}$.
\revised{Two objectives are defined to characterise the quality of a trajectory within each planning horizon.}

\textbf{Objective 1: Path Length.} Path length is minimised by penalising cumulative directional misalignment with the FMM vector field over the planning horizon:
$ f_1(\mathcal{T}_k) = \sum_{i=1}^{H} \left(1 - \frac{\mathbf{d}_i^k \cdot \mathbf{d}_{P,i}^k}{\|\mathbf{d}_i^k\| \, \|\mathbf{d}_{P,i}^k\|}\right) s(\mathbf{p}_i^k),
$
where \(\mathbf{d}_i^k = \mathbf{p}_i^k - \mathbf{p}_{i-1}^k\) is the displacement vector between consecutive states, \(\mathbf{d}_{P,i}^k\) is the ideal direction from the FMM field interpolated at position \(\mathbf{p}_i^k\) (terms with \(\|\mathbf{d}_i^k\| = 0\) contribute zero cost), and \(s(\mathbf{p})\) is a goal-proximity scaling factor:
$ s(\mathbf{p}) = 1 - \left( \frac{\|\mathbf{p} - \mathbf{p}_{\text{goal}}\|}{d_{\text{max}}} \right)^\mathfrak{s},$
where \(d_{\text{max}}\) is the maximum finite distance from vertices on the mesh to the goal and \(\mathfrak{s} > 1\) controls the rate of intensification near the goal as illustrated in Figure~\ref{fig:sp}. 

\textbf{Objective 2: Pose Stability.} Cumulative orientation violations relative to the local terrain are minimised to prevent tip-over. For each state, the terrain-aligned orientation is computed from the local terrain normal $\mathbf{n}_i^k = [n_x, n_y, n_z]^\top$ extracted from the neighbourhood descriptor $\mathfrak{d}_i^k$:
$ \theta_{\text{terrain},i}^k = \operatorname{atan2}\!\left(\nu_{\parallel},\, n_z\right), \quad \phi_{\text{terrain},i}^k = \operatorname{atan2}\!\left(-\nu_{\perp},\, \sqrt{\nu_{\parallel}^2 + n_z^2}\right),$ where $\nu_{\parallel} = -n_x \cos\psi_i^k - n_y \sin\psi_i^k$ and $\nu_{\perp} = n_x \sin\psi_i^k - n_y \cos\psi_i^k$,
defining the orientation perpendicular to the terrain at $\mathbf{p}_i^k$. The deviation from terrain-aligned orientation is:
$ \Delta\theta_i^k = \theta_i^k - \theta_{\text{terrain},i}^k, \quad \Delta\phi_i^k = \phi_i^k - \phi_{\text{terrain},i}^k.$
The stability objective penalises excessive deviations:
$ f_2(\mathcal{T}_k) = \sum_{i=0}^{H} \left[v_{\text{tilt}}(\Delta\theta_i^k) + v_{\text{tilt}}(\Delta\phi_i^k)\right],$
where the tilt violation function is:
$ v_{\text{tilt}}(\Delta\alpha) =
\begin{cases}
0, & \text{if } |\Delta\alpha| \leq \alpha_{\text{safe}} \\
\kappa\left(|\Delta\alpha| - \alpha_{\text{safe}}\right)^2, & \text{if } \alpha_{\text{safe}} < |\Delta\alpha| < \alpha_{\text{max}} \\
+\infty, & \text{if } |\Delta\alpha| \geq \alpha_{\text{max}}
\end{cases},$
with $\kappa > 0$ a penalty coefficient, $\alpha_{\text{safe}}$ the safe deviation angle from terrain-normal, and $\alpha_{\text{max}}$ the maximum allowable deviation before tip-over.

\section{Methodology}
We formulate uneven-terrain planning as a receding-horizon kinodynamic problem, seeking dynamically feasible trajectories that optimise multiple objectives under FMM global guidance. It is solved by $\epsilon$-constraint optimisation: at each iteration, forward trajectories are simulated, and the sequence minimising a progress objective is selected subject to a hard constraint whose tolerance is adjusted to local terrain. Candidates come from warm-starting and FMM-biased sampling, steering exploration toward promising regions. Along each trajectory, terrain descriptors are extracted, and orientation deviations from a terrain-aligned frame yield a physically meaningful stability-risk measure. The resulting controller adapts both its constraint threshold and sampling strategy to the terrain, spending less computation on benign terrain while automatically enforcing conservative behaviour in challenging conditions (Figure~\ref{fig:abstract}).

\subsection{Vehicle-Terrain Modelling}
\label{section:model}
Exact dynamics on uneven terrain are difficult to derive, so a nominal model is corrected by a learned residual, balancing fidelity with the fast computation required for real-time MPC.
With step size $\Delta t$, the nominal discrete-time dynamics are:
$\mathbf{x}_{k+1} =
\begin{bmatrix}
x_k + u_{0,k} \cos\theta_k \cos\psi_k \,\Delta t \\
y_k + u_{0,k} \cos\theta_k \sin\psi_k \,\Delta t \\
h(x_{k+1}, y_{k+1}) \\
\phi_{k+1} \\
\theta_{k+1} \\
\psi_k + u_{1,k} \,\Delta t
\end{bmatrix}
$
where $h(x,y)$ is the terrain height function. Roll and pitch follow from the terrain gradient rotated into the vehicle's heading frame. With central differences
$
\delta_x h(x, y) \approx \frac{h(x + \mathfrak{e},\, y) - h(x - \mathfrak{e},\, y)}{2\mathfrak{e}}, \quad
\delta_y h(x, y) \approx \frac{h(x,\, y + \mathfrak{e}) - h(x,\, y - \mathfrak{e})}{2\mathfrak{e}},
$
for small $\mathfrak{e} > 0$, the longitudinal and lateral slopes at $(x_{k+1}, y_{k+1})$ are
$
g_{\parallel} = \delta_x h \cos\psi_{k+1} + \delta_y h \sin\psi_{k+1}, \quad
g_{\perp} = -\delta_x h \sin\psi_{k+1} + \delta_y h \cos\psi_{k+1},
$
giving the pitch-then-roll attitude
$
\theta_{k+1} = \arctan g_{\parallel}, \quad
\phi_{k+1} = \operatorname{atan2}\!\left(-g_{\perp},\, \sqrt{1 + g_{\parallel}^2}\right).
$

\subsection{Terrain Height Function $h(x,y)$}
To find the terrain height at $(x,y)$, we select the mesh face $F^* \in \mathcal{M}$ whose projection onto the $xy$-plane contains $(x,y)$, with normal $\mathbf{n}=(a,b,c)$ and centre $\mathbf{p}_{\text{centre}}$, giving the plane $a x + b y + c z + d = 0$, where $d = -\mathbf{n}\cdot\mathbf{p}_{\text{centre}}$. The height is the vertical intersection of $(x,y)$ with this plane: $ h(x,y) = -\frac{a x + b y + d}{c}, \qquad c \neq 0. $

\subsection{Local Terrain Descriptor $\mathfrak{d}$}\label{sec:descriptor}
\begin{figure*}[!ht]
    \centering
    \includegraphics[width=0.6\linewidth]{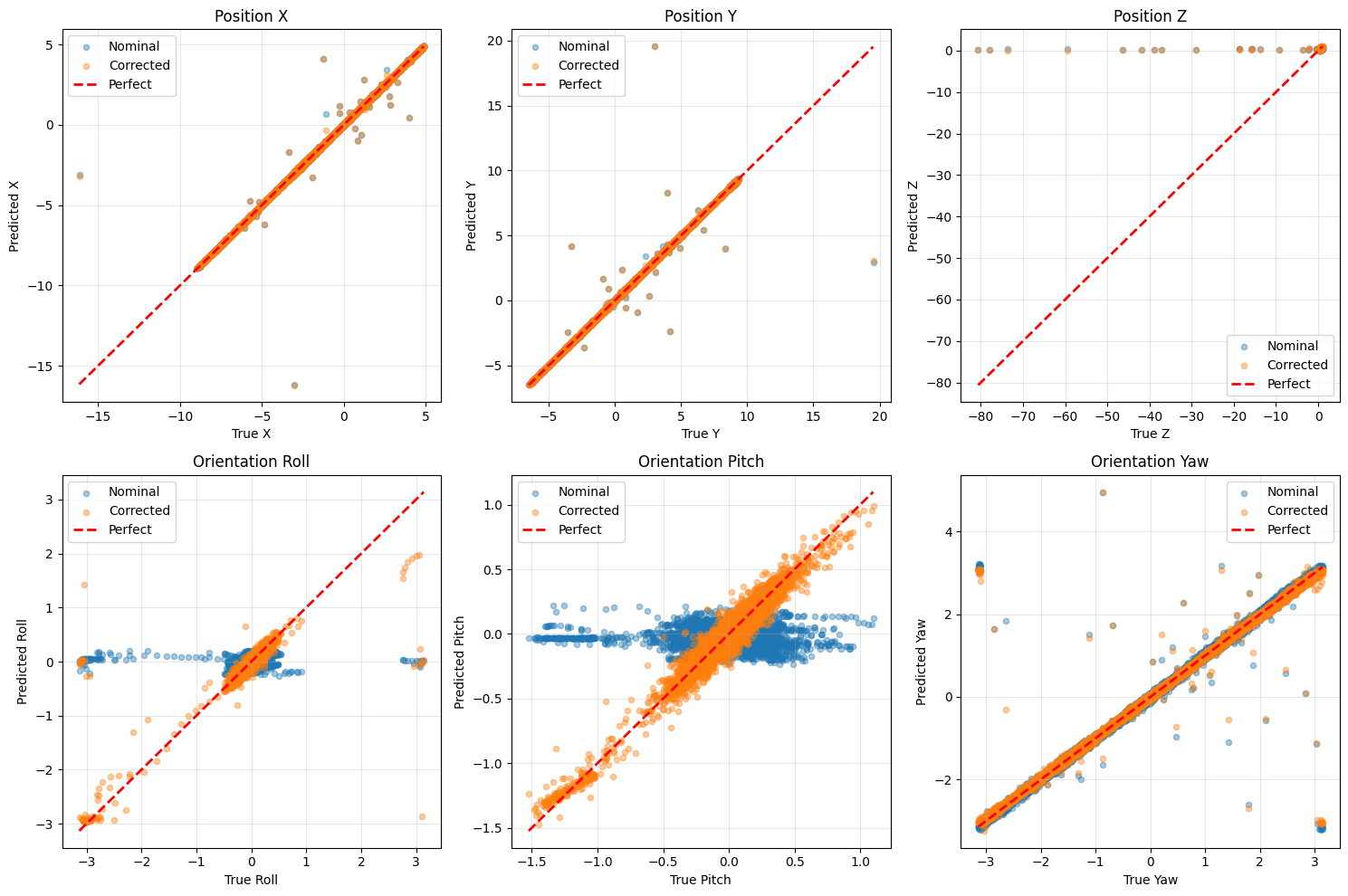}
    \caption{Comparison of nominal model, corrected model, and ground truth for robot pose estimation. While $x$, $y$, and yaw are well captured by the nominal model, roll and pitch predictions are (up to 32\%) improved by the Gaussian process residual correction; $z$ depends stochastically on terrain geometry.}
    \label{fig:model_correl}
\end{figure*}

To improve the accuracy of the proposed analytic model, we define a local terrain descriptor $\mathfrak{d} = [n_x, n_y, n_z, \sigma_z, K(\mathbf{q}), H(\mathbf{q}), A_{\text{total}}]$ that encodes geometric features of the terrain around the vehicle's footprint. The first three components are derived from the area- and distance-weighted average normal vector $\mathbf{n}_{\text{avg}} = {\sum_i w_i \mathbf{n}_i}/{\left\|\sum_i w_i \mathbf{n}_i\right\|}$, with weights $w_i = {A_i}/{(d_i + \mathfrak{e}_s)}$, where $\mathbf{n}_i$, $A_i$, and $d_i$ are the normal, area, and centroid distance of neighbouring triangle $T_i$, and $\mathfrak{e}_s$ is a small stabilising constant. Surface roughness is captured by the height variation $\sigma_z = \sqrt{{1}/{N} \sum_i (z_i - \bar{z})^2}$, where $z_i$ are centroid elevations within the local neighbourhood. For curvature, we extend the per-vertex operators of Meyer et al.~\cite{Meyer2003} to arbitrary query points $\mathbf{q}$ through distance-weighted interpolation $\omega_v = 1/(\|\mathbf{q} - \mathbf{p}_v\| + \mathfrak{e}_s)$ over vertices $v \in \mathfrak{N}(\mathbf{q})$
within a specified radius. The interpolated curvatures $K(\mathbf{q})$ and $H(\mathbf{q})$ are then computed as the weighted averages over neighbouring vertices, yielding continuous curvature fields queryable at arbitrary locations. The final component, $A_{\text{total}} = \sum_i A_i$, captures the
neighbourhood extent as the total area of triangles within the local radius.

\subsection{Semi-parametric Vehicle-Terrain Modelling}
While the nominal model provides a baseline prediction, unmodelled effects of the terrain can cause deviations. We model this deviation as a learned residual using Sparse Gaussian Process (SGP) regression, with one independent SGP per state dimension. Let the predicted state from the nominal model be \( \hat{\mathbf{x}}_{k+1} \), and the true observed state be \( \mathbf{x}_{k+1}^{\text{true}} \). The residual is: $\Delta \mathbf{x}_k = \mathbf{x}_{k+1}^{\text{true}} - \hat{\mathbf{x}}_{k+1}$. We model the residual as:
$\Delta \mathbf{x}_k = \mathbf{f}_{\text{res}}(\mathbf{x}_k, \mathbf{u}_k, \mathfrak{d}_k) + \boldsymbol{\varepsilon},
$ where $\boldsymbol{\varepsilon}$ is zero-mean Gaussian noise with covariance $\Sigma$. The SGP output is the 6D residual: $ \Delta \mathbf{x}_k = [\Delta x, \Delta y, \Delta z, \Delta \phi, \Delta \theta, \Delta \psi]^\top
$. The corrected next state is obtained as: $ \tilde{\mathbf{x}}_{k+1} = \hat{\mathbf{x}}_{k+1} + \Delta \mathbf{x}_k $. This residual correction improves prediction accuracy by capturing non-linear terrain-induced effects learned from data, as shown in Figure~\ref{fig:model_correl}.
Training data comprises ${\sim}5\times 10^{4}$ tuples $(\mathbf{x}_k, \mathbf{u}_k, \mathfrak{d}_k, \mathbf{x}_{k+1}^{\text{true}})$ from exploratory rollouts in simulation and on the physical Scout Mini over both meshes, with a held-out subset reserved for the evaluation in Figure~\ref{fig:model_correl}. The SGP uses the variational free-energy objective of Titsias~\cite{titsias2009variational} ($M=500$ inducing points) with a Mat\'ern--$5/2$ ARD kernel over the $15$-dimensional input $(\mathbf{x}, \mathbf{u}, \mathfrak{d})$; inducing locations, kernel hyperparameters, and noise covariance $\Sigma$ are optimised jointly by maximising the variational lower bound. Six independent SGPs model the six residual components. At deployment, only the predictive mean is queried, giving $O(M)$ cost per output dimension and keeping the correction within the per-iteration planning time budget.
\subsection{Solution By Terrain-Adaptive $\epsilon$-Constraint Optimisation}
\begin{algorithm}
\caption{Terrain-Adaptive $\epsilon$-Constraint Kinodynamic Planning}\label{alg:planner}
\begin{algorithmic}[1]
\REQUIRE Current state $x_0$, base stability budget $\epsilon_{\text{base}}$, terrain descriptor $\mathfrak{d}_0$, number of candidates $N_{\text{cand}}$, horizon $H$, FMM bias $\alpha_{\text{FMM}}$
\STATE Compute adaptive epsilon: $\epsilon_{\text{stab}} \gets \epsilon_{\text{stab}}(\mathfrak{d}_0)$
\STATE prev\_best\_controls $\gets$ optional previous solution
\FOR{$j = 1$ \TO $N_{\text{cand}}$}
    \IF{$j \le N_{\text{cand}}/2$ \AND prev\_best\_controls available}
        \STATE Generate $U^{(j)}$ via warm start: shift and add noise
    \ELSE
        \STATE Generate $U^{(j)}$ and $\mathcal{T}^{(j)}$ via FMM-biased sampling:
        \STATE \quad Roll out from $x_0$, blending FMM-suggested and random controls at each step
    \ENDIF
    \STATE \textbf{if} $\mathcal{T}^{(j)}$ not yet computed: $\mathcal{T}^{(j)} \gets \text{simulate\_forward}(x_0, U^{(j)})$
    \STATE Extract terrain descriptors: $\{\mathfrak{d}_i^{(j)}\}_{i=0}^H$ along $\mathcal{T}^{(j)}$
    \STATE Compute terrain-aligned orientations: $\{\theta_{\text{terrain},i}^{(j)}, \phi_{\text{terrain},i}^{(j)}\}_{i=0}^H$
    \STATE Compute orientation deviations: $\{\Delta\theta_i^{(j)}, \Delta\phi_i^{(j)}\}_{i=0}^H$
    \STATE Compute objectives: $f_1^{(j)} \gets f_1(\mathcal{T}^{(j)})$, $f_2^{(j)} \gets f_2(\mathcal{T}^{(j)})$
\ENDFOR
\STATE feasible $\gets$ $\{ j \mid f_2^{(j)} \le \epsilon_{\text{stab}} \text{ and no hard violations} \}$
\IF{feasible is not empty}
    \STATE best $\gets$ $\arg\min_{j \in \text{feasible}} f_1^{(j)}$
    \STATE Apply first control: $\mathbf{u}_0 \gets U^{(\text{best})}[0]$
    \STATE Store previous controls: prev\_best\_controls $\gets U^{(\text{best})}$
\ELSE 
\STATE Perform safe fallback manoeuvre
\ENDIF
\end{algorithmic}
\end{algorithm} 
At each planning iteration \(k\), we employ the terrain-adaptive $\epsilon$-constraint method (Algorithm~\ref{alg:planner}) to transform the bi-objective problem into a constrained single-objective optimisation over the horizon \(H\). $\epsilon_{\text{stab}}$ is dynamically computed based on the current terrain characteristics at the robot's position. This adaptation makes the planner more conservative in challenging terrain while permitting greater exploration in benign conditions.
At each planning iteration $k$, we solve the problem:
$\begin{aligned}
    \min_{\mathcal{T}_k} \quad & f_1(\mathcal{T}_k) \\
    \text{subject to} \quad & \mathbf{x}_{i+1}^k = f(\mathbf{x}_i^k, \mathbf{u}_i^k, \mathfrak{d}_i^k, \Delta t), \quad i = 0, \ldots, H-1, \\
    & f_2(\mathcal{T}_k) \leq \epsilon_{\text{stab}}(\mathfrak{d}_k), \\
    & \mathbf{x}_0^k = \mathbf{x}_k, \\
    & \mathbf{p}_i^k \in \mathcal{F}, \quad i = 0, \ldots, H, \\
    & \mathbf{u}_i^k \in \mathcal{U}, \quad i = 0, \ldots, H-1,
\end{aligned}
$
where \(\mathcal{F}\) represents the collision-free navigation space on the mesh, \(\mathcal{U}\) represents the admissible control set, and \(\epsilon_{\text{stab}}(\mathfrak{d}_k)\) is the terrain-adaptive maximum allowable cumulative stability violation over the planning horizon, computed based on the current terrain descriptor \(\mathfrak{d}_k\) at the robot's position. The stability constraint threshold \(\epsilon_{\text{stab}}\) adapts to local terrain difficulty through the following formulation: $\epsilon_{\text{stab}}(\mathfrak{d}) = \epsilon_{\text{base}} \cdot \max\!\left(0.3,\, \min(f_r(\mathfrak{d}), f_s(\mathfrak{d}))\right),$ where the reduction factors based on terrain roughness and slope are computed as: $
\begin{aligned}
    f_r(\mathfrak{d}) &= 1.0 - \gamma_r \cdot \min\left(1.0, \frac{\mathfrak{d}.\text{roughness}}{\mathfrak{d}_{\max}.\text{roughness}}\right), \\
    f_s(\mathfrak{d}) &= 1.0 - \gamma_s \cdot \min\left(1.0, \frac{\mathfrak{d}.\text{inclination}}{\pi/2}\right),
\end{aligned}
$ where \(\gamma_r\) and \(\gamma_s\) are adaptation strength parameters controlling the sensitivity to roughness and slope, respectively, $\mathfrak{d}.\text{roughness} \equiv \sigma_z$ and $\mathfrak{d}.\text{inclination} \equiv \arccos(n_z)$.
The combined factor is lower-bounded at 0.3 to prevent overly restrictive planning. More challenging terrain (higher roughness or steeper slopes) yields smaller reduction factors, thereby tightening the stability budget and enforcing more conservative trajectory selection. The scaling functions and combination rule implement a worst-feature-dominates design. Roughness and inclination scale linearly, the simplest monotone form, as neither descriptor has a known threshold justifying a higher-order response. Normalising by $\text{roughness}_{\max}$ and $\pi/2$ maps both to $[0,1]$, giving $\gamma_r$ and $\gamma_s$ a common reading as fractional budget reduction at worst-case input. Combining via $\min(f_r, f_s)$, rather than a product or weighted sum, prevents a hazardous roughness or slope value from being averaged away by a benign one. The floor at $0.3$ keeps $\epsilon_{\text{stab}}$ from collapsing on sharp-feature transients. We solve the constrained optimisation problem using a sampling-based approach that generates \(N_{\text{cand}}\) candidate control sequences and selects the best feasible solution. To leverage the global path guidance from the FMM vector field, we employ a hybrid sampling strategy that combines warm-starting from previous solutions with FMM-biased exploration. 

\textbf{Candidate Generation:} Two complementary strategies generate the control sequences.
\textit{Strategy 1: Warm Start.} When a previous solution exists, half the candidates shift the previous best control sequence forward one step and add Gaussian noise, providing temporal continuity and faster convergence: $\mathbf{u}_i^{(j)} = \mathbf{u}_{i+1}^{\text{prev}} + \boldsymbol{\eta}_i$ for $i = 0,\ldots,H-2$, with the final control repeated, $\mathbf{u}_{H-1}^{(j)} = \mathbf{u}_{H-1}^{\text{prev}} + \boldsymbol{\eta}_{H-1}$, where $\boldsymbol{\eta}_i \sim \mathcal{N}(0, \boldsymbol{\Sigma}_{\text{noise}})$, with $\boldsymbol{\Sigma}_{\text{noise}} = \text{diag}(\sigma_v^2, \sigma_\omega^2)$ controlling exploration around it.
\textit{Strategy 2: FMM-Biased Random Sampling.} The remaining candidates blend FMM-suggested controls with random exploration. Simulating forward, at each state $\mathbf{x}_i$ we query the FMM field $\mathbf{d}_{\text{FMM}}(\mathbf{p}_i)$ for a desired heading $\psi_{\text{des},i} = \operatorname{atan2}(d_{\text{FMM},y}, d_{\text{FMM},x})$, then form the FMM-suggested control via a proportional controller, $\mathbf{u}_{\text{FMM},i} = \begin{bmatrix} \beta v_{\max} \\ k_p(\psi_{\text{des},i} - \psi_i) \end{bmatrix}$, where $\beta \in (0,1]$ scales velocity conservatively, $k_p$ is a proportional gain, and the heading error $\psi_{\text{des},i} - \psi_i$ is wrapped to $[-\pi, \pi]$. This is blended with a random control $\mathbf{u}_{\text{rand},i}$ sampled uniformly from $\mathcal{U}$: $\mathbf{u}_i^{(j)} = \alpha_{\text{FMM}} \mathbf{u}_{\text{FMM},i} + (1 - \alpha_{\text{FMM}}) \mathbf{u}_{\text{rand},i}$, where $\alpha_{\text{FMM}} \in [0,1]$ sets the bias strength: $\alpha_{\text{FMM}} = 0$ gives pure random sampling, $\alpha_{\text{FMM}} = 1$ strictly follows the FMM field; all candidate controls are clipped component-wise to $\mathcal{U}$.

The terrain-adaptive pose stability formulation inherently adapts to local geometry: orientations unstable on flat terrain become necessary on moderate slopes, while excessively steep regions are marked as lethal by the FMM, whose vector field points the vehicle away. The terrain-adaptive $\epsilon$-constraint reinforces this, automatically tightening stability requirements in challenging terrain while permitting more aggressive exploration in benign conditions.

\section{Evaluation}
We evaluate the proposed terrain-adaptive planner in simulation across diverse terrain, focusing on three questions: (1) Does FMM-biased sampling improve trajectory quality over pure random sampling? (2) How does the terrain-adaptive $\epsilon$-constraint affect stability and path efficiency across terrain difficulties? (3) Can the planner replan in real time while maintaining safety?

\subsection{Experimental Setup}
Experiments use ROS 2 Humble with Gazebo on two 3D meshes: Mesh 1 (270,904 vertices, 536,879 faces) and Mesh 2 (59,107 vertices, 117,806 faces). Fifty start-end positions per mesh, labelled rough or gentle-slope, are each run 5 times to account for stochastic sampling. We simulate an AgileX Scout Mini using the model in Section~\ref{section:model} with $v_{\max} = 1.5$\,m/s, $\omega_{\max} = 3.22$\,rad/s, safe orientation threshold $\alpha_{\text{safe}} = 30^{\circ}$, and penalty coefficient $\kappa = 10.0$. Planning parameters include horizon $H = 10$ steps ($1.0$\,s), control timestep $\Delta t = 0.1$\,s, $N_{\text{cand}} = 100$ candidate trajectories, FMM bias $\alpha_{\text{FMM}} = 0.7$, base stability budget $\epsilon_{\text{base}} = 0.5$, terrain adaptation strengths $\gamma_r = 0.6$ and $\gamma_s = 0.7$, and noise standard deviations $\sigma_v = 0.5$\,m/s and $\sigma_\omega = 0.3$\,rad/s. We compare against two ablation baselines: pure random sampling ($\alpha_{\text{FMM}} = 0$) and fixed $\epsilon$ (no terrain adaptation). Five metrics are reported. Path-length deviation $\Delta_L = L - L_g$ is the excess of the executed path length $L$ over the FMM geodesic length $L_g$ between start and goal. Maximum orientation deviation $\phi_{\max} = \max_i |\phi_i|$ is the peak roll magnitude relative to the gravity-aligned horizontal along the trajectory. The trade-off $T = 0.5\,\bar{\phi} + 0.5\,\bar{\Delta}_L$ (lower is better) uses fixed normalisations $\bar{\phi} = \phi_{\max}/\alpha_{\max}$ and $\bar{\Delta}_L = \Delta_L/L_g$, computed per trial. Average planning time is the mean solver time per receding-horizon iteration, and success rate is the percentage of trials in which the vehicle reached the goal ($\|\mathbf{p}-\mathbf{p}_{\text{goal}}\| < \delta_{\text{term}}$) without a tip-over or solver failure. Setup available at \url{https://github.com/Stblacq/mesh_planner}.

\begin{table}[htbp]
\centering
\caption{Effect of FMM-Biased Sampling on Navigation Performance}
\label{tab:fmm_ablation}
\resizebox{0.8\linewidth}{!}{%
\begin{tabular}{llccccc}
\toprule
\textbf{Terrain Type} & \textbf{$\alpha_{\text{FMM}}$} & \textbf{Success} & \textbf{Path Length} & \textbf{Max Orient.} & \textbf{Trade-off} & \textbf{Planning} \\
 & & \textbf{Rate (\%)} & \textbf{Dev. $\Delta_L$ (m)} & \textbf{Dev. $\phi_{\text{max}}$ (°)} & \textbf{Metric $T$} & \textbf{Time (ms)} \\
\midrule
\multirow{3}{*}{Gentle Slopes} 
  & 0.0 & 87 & 8.6 & 15.3 & 0.62 & 38.2 \\
  & 0.7 & \textbf{98} & \textbf{4.2} & 13.8 & \textbf{0.35} & 42.1 \\
  & 1.0 & 94 & 5.1 & 14.1 & 0.41 & 40.5 \\
\midrule
\multirow{3}{*}{Rough Terrain} 
  & 0.0 & 62 & 15.4 & 28.4 & 0.78 & 45.6 \\
  & 0.7 & \textbf{89} & \textbf{9.8} & 26.7 & \textbf{0.54} & 48.3 \\
  & 1.0 & 78 & 11.2 & 29.2 & 0.63 & 46.9 \\
\bottomrule
\end{tabular}
}
\end{table}

\begin{table}[htbp]
\centering
\caption{Planning Horizon Sensitivity}
\label{tab:horizon_sensitivity}
\resizebox{0.8\linewidth}{!}{%
\begin{tabular}{lcccc}
\toprule
\textbf{Horizon $H$} & \textbf{Success} & \textbf{Path Length} & \textbf{Trade-off} & \textbf{Planning} \\
\textbf{(steps)} & \textbf{Rate (\%)} & \textbf{Dev. $\Delta_L$ (m)} & \textbf{Metric $T$} & \textbf{Time (ms)} \\
\midrule
5   & 82 & 14.2 & 0.72 & 12 \\
10  & 90 & 12.8 & 0.67 & 18 \\
15  & 91 & 11.9 & 0.64 & 28 \\
20  & 94 & 11.2 & 0.61 & 42 \\
25  & 95 & 10.8 & 0.59 & 56 \\
\bottomrule
\end{tabular}
}
\end{table}

\begin{table}[htbp]
\centering
\caption{Candidate Sample Size Sensitivity}
\label{tab:sample_sensitivity}
\resizebox{0.8\linewidth}{!}{%
\begin{tabular}{lcccc}
\toprule
\textbf{$N_{\text{cand}}$} & \textbf{Success} & \textbf{Path Length} & \textbf{Trade-off} & \textbf{Planning} \\
 & \textbf{Rate (\%)} & \textbf{Dev. $\Delta_L$ (m)} & \textbf{Metric $T$} & \textbf{Time (ms)} \\
\midrule
50   & 78 & 13.8 & 0.76 & 18 \\
75   & 85 & 11.5 & 0.65 & 25 \\
80   & 88 & 10.8 & 0.61 & 28 \\
100  & 92 & 9.5 & 0.55 & 32 \\
120  & 95 & 8.7 & 0.51 & 38 \\
\bottomrule
\end{tabular}
}
\end{table}

\begin{table}[htbp]
\centering
\caption{Terrain Adaptation Parameter Sensitivity}
\label{tab:adaptation_sensitivity}
\resizebox{0.8\linewidth}{!}{%
\begin{tabular}{cccccc}
\toprule
\textbf{$\gamma_r$} & \textbf{$\gamma_s$} & \textbf{Success} & \textbf{Path Length} & \textbf{Max Orient.} & \textbf{Trade-off} \\
 & & \textbf{Rate (\%)} & \textbf{Dev. $\Delta_L$ (m)} & \textbf{Dev. $\phi_{\text{max}}$ (°)} & \textbf{Metric $T$} \\
\midrule
0.4 & 0.4 & 89 & 7.2 & 31.5 & 0.63 \\
0.5 & 0.5 & 92 & 7.8 & 27.3 & 0.56 \\
0.6 & 0.7 & 96 & 8.5 & 22.4 & 0.50 \\
0.7 & 0.8 & 98 & 9.4 & 18.2 & 0.47 \\
0.8 & 0.8 & 100 & 10.8 & 14.5 & 0.45 \\
\bottomrule
\end{tabular}
}
\end{table}

\begin{table}[htbp]
\centering
\caption{$\epsilon$-Constraint: Terrain-Adaptive vs. Fixed}
\label{tab:epsilon_comparison}
\resizebox{0.8\linewidth}{!}{%
\begin{tabular}{lccccc}
\toprule
\textbf{Method} & \textbf{Success} & \textbf{Path Length} & \textbf{Max Orient.} & \textbf{Trade-off} & \textbf{Tip-over} \\
 & \textbf{Rate (\%)} & \textbf{Dev. $\Delta_L$ (m)} & \textbf{Dev. $\phi_{\text{max}}$ (°)} & \textbf{Metric $T$} & \textbf{Incidents} \\
\midrule
Fixed $\epsilon$ ($\epsilon_{\text{stab}} = 0.25$) & 83 & 9.2 & 29.8 & 0.61 & 2 \\
Terrain-Adaptive & \textbf{94} & \textbf{8.5} & \textbf{20.3} & \textbf{0.48} & \textbf{0} \\
\midrule
\textit{Improvement} & +13\% & +7.6\% & +32\% & +21\% & +100\% \\
\bottomrule
\end{tabular}
}
\end{table}
\subsection{Ablation Studies}
\textbf{Effect of FMM-Biased Sampling.} Table~\ref{tab:fmm_ablation} shows that FMM-biased sampling at $\alpha_{\text{FMM}} = 0.7$ improves all trajectory-quality metrics on both terrain types, giving the largest gains in success rate, path-length deviation, and trade-off $T$. Full bias ($\alpha_{\text{FMM}} = 1.0$) is consistently worse than $0.7$ on these quality metrics, indicating that balanced exploration-exploitation is needed to escape local minima on complex terrain, while the associated increase in planning time is minor relative to these gains. 

\textbf{Terrain-Adaptive vs.\ Fixed $\epsilon$-Constraint.} Table~\ref{tab:epsilon_comparison} confirms the value of terrain-adaptive constraints: the adaptive scheme improves success rate, trade-off, and maximum orientation deviation over a fixed $\epsilon$, and most critically eliminates all tip-over incidents. By relaxing the stability budget on favourable terrain and tightening it on difficult sections, it also produces more direct paths.
\subsection{Sensitivity Analysis}
\textbf{Horizon Length.} Longer horizons improve success rate and trade-off $T$, with gains in $T$ diminishing, while planning time grows steeply with $H$ (Table~\ref{tab:horizon_sensitivity}). Gains beyond $H{=}10$ are marginal, yet planning time more than triples. We adopt $H{=}10$, trading minor quality for faster replanning.

\textbf{Candidate Sample Size.} Table~\ref{tab:sample_sensitivity} shows trajectory quality improving monotonically with $N_{\text{cand}}$ at an approximately linear time cost. Beyond $N_{\text{cand}} = 100$, the marginal gain per additional candidate diminishes relative to the linear growth in planning time; we therefore adopt $N_{\text{cand}} = 100$ as the practical operating point.

\textbf{Adaptation Parameters.} Table~\ref{tab:adaptation_sensitivity} shows that stronger adaptation ($\gamma_r,\gamma_s$) reduces maximum orientation deviation at the cost of longer paths. Although $T$ is minimised at $(0.8,0.8)$, gains beyond $(0.6,0.7)$ are marginal and bought with longer paths, so we retain $(0.6,0.7)$.

\subsection{Comparative Analysis}
\begin{table}[htbp]
\centering
\caption{Comparison with Baseline Methods}
\label{tab:baseline_comparison}
\resizebox{0.8\linewidth}{!}{%
\begin{tabular}{lccccc}
\toprule
\textbf{Method} & \textbf{Success} & \textbf{Path Length} & \textbf{Max Orient.} & \textbf{Trade-off} & \textbf{Planning} \\
 & \textbf{Rate (\%)} & \textbf{Dev. $\Delta_L$ (m)} & \textbf{Dev. $\phi_{\text{max}}$ (°)} & \textbf{Metric $T$} & \textbf{Time (ms)} \\
\midrule
MPPI & 87 & \textbf{7.8} & 24.6 & 0.62 & \textbf{29} \\
GAKD & 89 & 9.2 & 19.5 & 0.56 & 38 \\
$\epsilon$-MPC (Proposed) & \textbf{94} & 8.5 & \textbf{18.7} & \textbf{0.48} & 35 \\
\midrule
\textit{Improvement vs. MPPI} & +8\% & -9\% & +24\% & +23\% & -21\% \\
\textit{Improvement vs. GAKD} & +6\% & +8\% & +4\% & +14\% & +8\% \\
\bottomrule
\end{tabular}
}
\end{table}
Table~\ref{tab:baseline_comparison} compares the proposed terrain-adaptive FMM-biased approach against two sampling-based baselines, MPPI and GAKD, both employing weighted scalarisation of path length and stability costs but lacking FMM guidance and terrain-adaptive constraints. The proposed method achieves a 94\% success rate with trade-off metric $T = 0.48$, outperforming MPPI (87\%, $T = 0.62$) and GAKD (89\%, $T = 0.56$), with particularly robust handling of slope scenarios where baselines struggle with local minima. Computationally, the proposed approach requires 35\,ms per iteration, a 21\% increase over MPPI (29\,ms) but 8\% faster than GAKD (38\,ms), which is within real-time constraints. While MPPI produces the shortest paths ($\Delta_L = 7.8$\,m vs.\ 8.5\,m proposed, 9.2\,m GAKD), the proposed method achieves the best stability with $\phi_{\text{max}} = 18.7^{\circ}$ compared to GAKD's $19.5^{\circ}$ and MPPI's $24.6^{\circ}$, effectively sacrificing 9\% path efficiency relative to MPPI while gaining 8\% and 6\% higher success rates over MPPI and GAKD, respectively.
\subsection{Real-World Experiments}
\label{sec:realworld}
\begin{figure}[htbp]
    \centering
    \includegraphics[width=\linewidth]{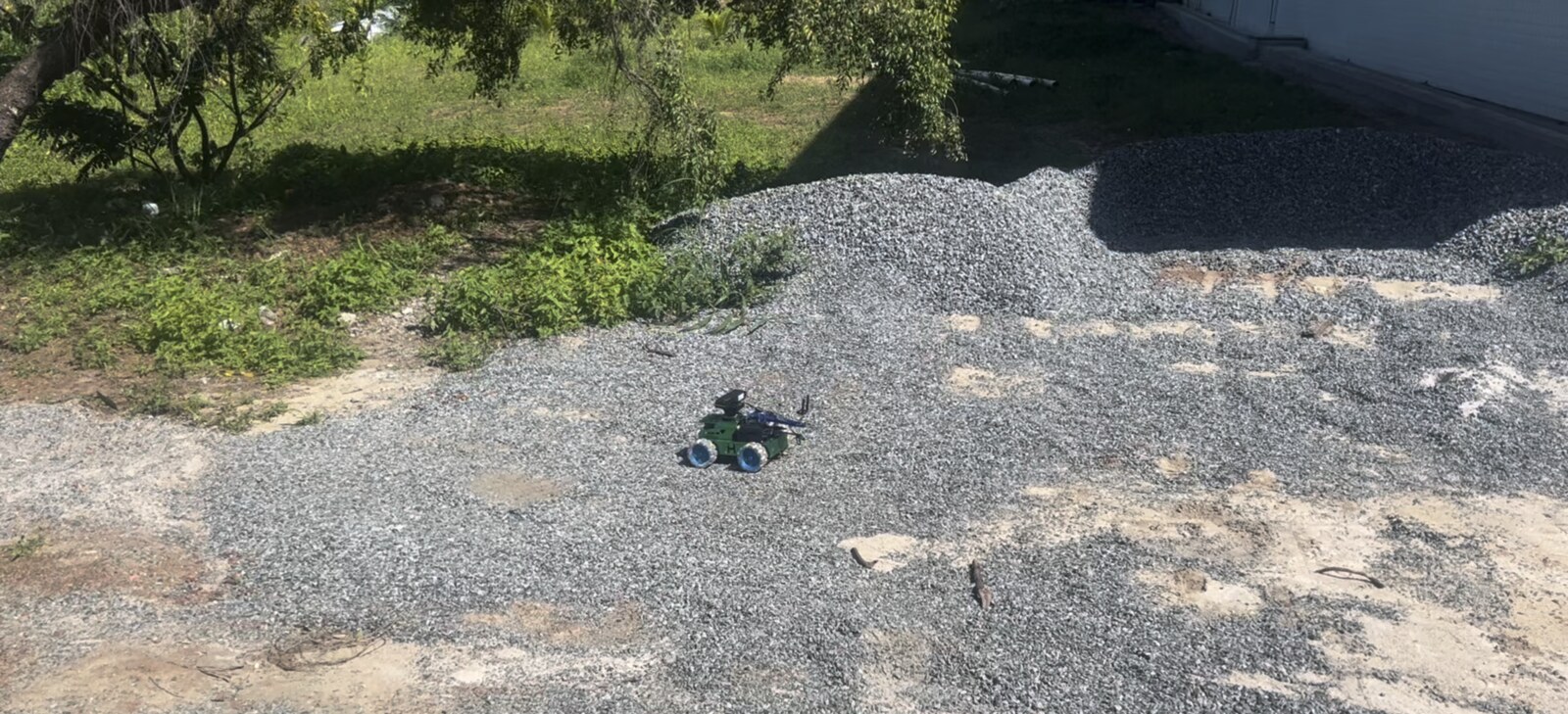}
    \caption{\revised{Real-world deployment of the platform. The JetAuto robot operating on uneven gravel terrain during outdoor field trials.}}
    \label{fig:real_snapshots}
\end{figure}

To validate the simulation findings, we deployed the planner on a physical AgileX Scout Mini outdoors, equipped with a Velodyne VLP-16 LiDAR for terrain perception and a workstation (Intel Core i7, 32\,GB RAM, NVIDIA RTX 4050 GPU, Ubuntu Linux). Real-world results follow the simulation trends: success rate drops from $94\%$ to $80\%$ due to unmodelled effects such as wheel slip and sensor noise, while path-length deviation rises from $8.5$\,m to $10.8$\,m as the planner takes more conservative routes under uncertainty. Planning times rise from $35$\,ms to $48.7$\,ms but still meet the real-time requirement. The method's advantages over the baselines persist in physical deployment, confirming that terrain-adaptive constraints scale to real-world conditions.

\begin{table}[htbp]
\centering
\caption{Real-World Baseline Comparison}
\label{tab:realworld_baseline}
\resizebox{0.8\linewidth}{!}{%
\begin{tabular}{lccccc}
\toprule
\textbf{Method} & \textbf{Success} & \textbf{Path Length} & \textbf{Max Orient.} & \textbf{Trade-off} & \textbf{Planning} \\
 & \textbf{Rate (\%)} & \textbf{Dev. $\Delta_L$ (m)} & \textbf{Dev. $\phi_{\text{max}}$ (°)} & \textbf{Metric $T$} & \textbf{Time (ms)} \\
\midrule
MPPI & 67 & \textbf{10.5} & 32.8 & 0.71 & 42.3 \\
GAKD & 73 & 13.1 & 30.2 & 0.65 & 51.6 \\
$\epsilon$-MPC (Proposed) & \textbf{80} & 10.8 & \textbf{26.4} & \textbf{0.58} & 48.7 \\
\bottomrule
\end{tabular}
}
\end{table}

Table~\ref{tab:realworld_baseline} shows that the proposed method achieves an 80\% success rate compared to MPPI's 67\% and GAKD's 73\%. The stability improvements are particularly pronounced, with the proposed method limiting orientation deviation to $26.4^{\circ}$ compared to GAKD's $30.2^{\circ}$ and MPPI's $32.8^{\circ}$, maintaining a safe margin from the vehicle's $35^{\circ}$ mechanical tip-over threshold. MPPI's failures resulted from excessive roll angles on steep terrain sections, while the proposed terrain-adaptive approach successfully navigated these same regions by tightening stability constraints based on local slope measurements.

\revised{\textbf{JetAuto Platform Validation.}} To strengthen the real-world evaluation, the planner is also deployed
on a four-wheeled JetAuto mobile base (Figure~\ref{fig:real_snapshots}). The SGP is retrained on a fresh dataset of nominal-prediction errors collected on the new platform. Experiments are conducted on an outdoor construction site with uneven terrain using fifty start-goal pairs per method
(Table~\ref{tab:second-platform_baseline}).
\begin{table}[htbp]
\centering
\caption{JetAuto: Outdoor Off-Road Baseline Comparison}
\label{tab:second-platform_baseline}
\resizebox{0.8\linewidth}{!}{%
\begin{tabular}{lccccc}
\toprule
\textbf{Method} & \textbf{Success} & \textbf{Path Length} & \textbf{Max Orient.} & \textbf{Trade-off} & \textbf{Planning} \\
 & \textbf{Rate (\%)} & \textbf{Dev. $\Delta_L$ (m)} & \textbf{Dev. $\phi_{\text{max}}$ (°)} & \textbf{Metric $T$} & \textbf{Time (ms)} \\
\midrule
MPPI                       & 58 & 12.4 & 35.7 & 0.78 & 44.1 \\
GAKD                       & 66 & 14.8 & 32.5 & 0.70 & 53.2 \\
$\epsilon$-MPC (Proposed)  & \textbf{74} & \textbf{11.9} & \textbf{28.6} & \textbf{0.62} & 50.4 \\
\bottomrule
\end{tabular}
}
\end{table}
The trends from the Scout Mini and simulation evaluations persist on
this platform and the harder terrain: the $\epsilon$-MPC retains a
better trade-off while staying within the real-time planning budget.

\subsection{Failure Analysis}
Failed trials showed three modes: infeasible goals in FMM-unreachable regions, local-minima trapping in narrow corridors (mitigable by increasing $N_{\text{cand}}$), and stability-constraint violations at terrain discontinuities where descriptor extraction incompletely captures local geometry. We also tried NSGA-II to jointly optimise path efficiency and stability, but its population-based search exceeded 2000\,ms per iteration, over an order of magnitude slower than MPPI (29\,ms) and GAKD (38\,ms), while yielding comparable or worse success rates, confirming that for real-time receding-horizon planning, simpler sampling-based methods with direct FMM guidance are more efficient without sacrificing quality.
\section{Conclusion and Future Work}
While weighted scalarisation can optimise individual objectives effectively, the proposed terrain-adaptive $\epsilon$-constraint better balances \revised{the two objectives} by dynamically adjusting stability constraints from local terrain features, enabling safer navigation without excessive path deviation. Its modest computational overhead is justified by substantial gains in feasibility and completion rate, particularly critical for autonomous systems in unstructured outdoor environments. The approach thus provides a principled framework balancing global optimality, local safety, and computational tractability. Future directions include extending it to dynamic environments, integrating temporal prediction into FMM computation and stability assessment for navigation among moving obstacles, and enriching the terrain descriptor with semantic properties such as surface softness to refine traversability assessment and broaden deployable environments.

\bibliographystyle{IEEEtran}
\bibliography{ref} 
\end{document}